\documentclass[10pt,twocolumn,letterpaper]{article}

\usepackage[pagenumbers]{wacv} 

\usepackage[accsupp]{axessibility}  

\graphicspath{{./fig/}}

\definecolor{wacvblue}{rgb}{0.21,0.49,0.74}
\usepackage[pagebackref,breaklinks,colorlinks,allcolors=wacvblue]{hyperref}

\usepackage{algorithm}
\usepackage{algpseudocode}
\definecolor{comment}{rgb}{0,0.6,0.3}

\def\wacvPaperID{VisionDocs-16} 
\def\confName{WACV}
\def\confYear{2026}

\title{Pseudo Contrastive Learning\\for Diagram Comprehension in Multimodal Models}

\author{Hiroshi Sasaki\\
\href{https://www.jri.co.jp/en/}{The Japan Research Institute, Limited}\\
{\tt\small \href{mailto:sasaki.hiroshi@jri.co.jp}{\url{sasaki.hiroshi@jri.co.jp}}}
}

\begin{document}
\maketitle
\begin{abstract}
Recent multimodal models such as Contrastive Language-Image Pre-training (CLIP) have shown remarkable ability to align visual and linguistic representations. However, domains where small visual differences carry large semantic significance, such as diagram understanding, remain challenging due to the models' limited sensitivity to fine-grained structural variations.

We propose a new training paradigm designed to enhance diagram comprehension in vision-language models. Our approach introduces pseudo contrastive samples generated by a diagram renderer that creates synthetic diagrams using randomly picked text elements. These samples highlight structural differences in diagrammatic imagery without requiring any modification or editing of the original data. By incorporating these pseudo contrastive samples into the training objective, the model learns to capture more precise and semantically consistent diagram structures.

Empirical evaluations on a benchmark dataset of flowcharts demonstrate substantial improvements over standard CLIP and hard-negative CLIP training in both image-text matching and visual question answering tasks. The results underscore the value of domain-specific training strategies and contribute to advancing diagrammatic understanding within the broader context of vision-language learning.
\end{abstract}    
\section{Introduction}
\label{sec:intro}

\begin{figure}[t]
    \centering
    \includegraphics[width=1.0\linewidth]{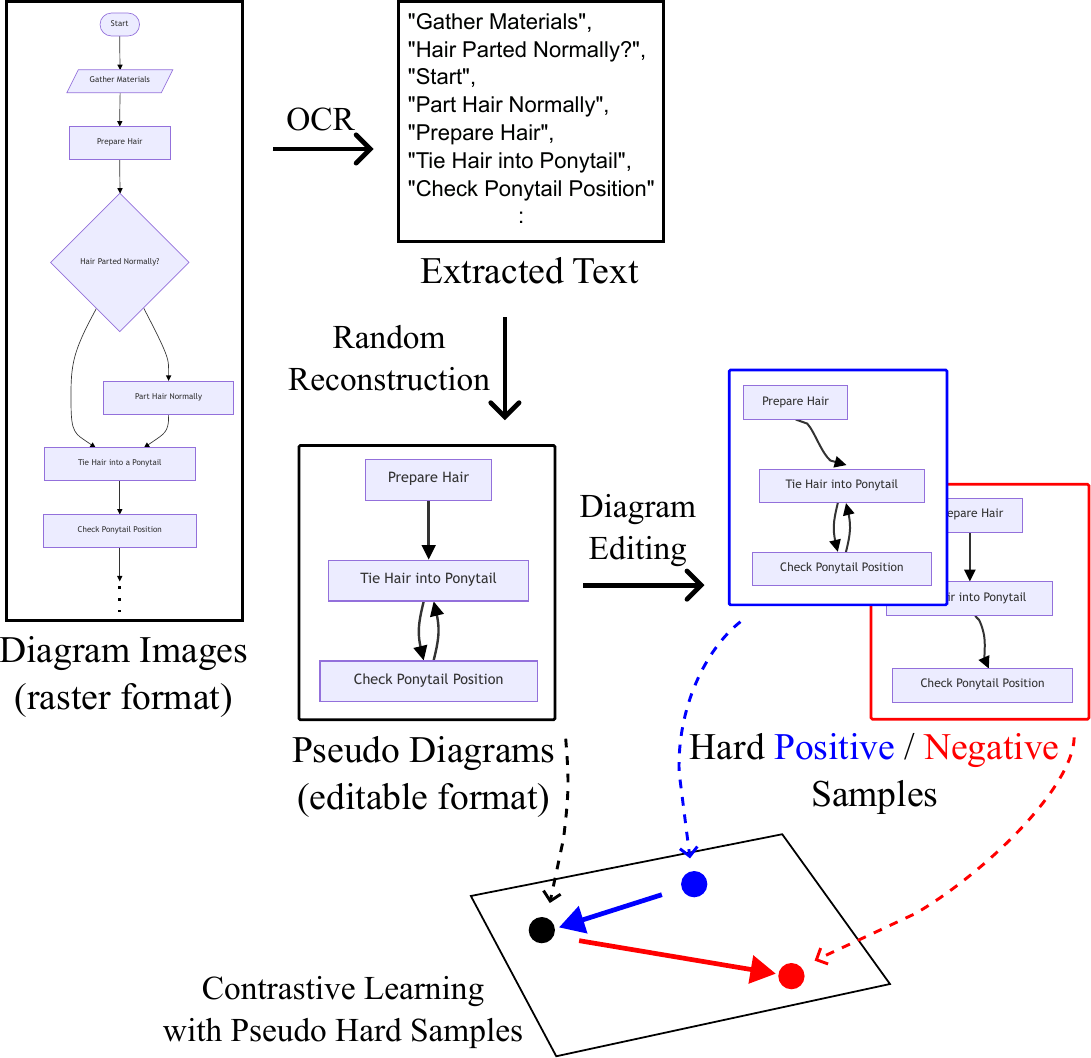}
    \caption{Conceptual overview of our proposed method. Text is extracted from a raster image via OCR to generate pseudo diagrams with random connections. These are rendered into an editable format, from which hard positive and negative pairs are created through rule-based edits. The resulting samples are used to train a VLM with structure-aware contrastive learning, enhancing its ability to distinguish fine-grained structural differences.}
    \label{fig:overview}
\end{figure}

Vision-Language Models (VLMs), such as CLIP (Contrastive Language-Image Pre-training)~\cite{radford2021learning}, have emerged as powerful tools, integrating visual and linguistic information to achieve remarkable capabilities in image understanding. These models have demonstrated superior performance across diverse tasks, ranging from image captioning and visual question answering (VQA) to zero-shot recognition, by effectively bridging the modalities of visual perception and language. Their ability to link visual and textual inputs has advanced artificial intelligence applications.

Meanwhile, diagrams, exemplified by flowcharts, serve as indispensable visual representations for conveying complex concepts and intricate procedures that are challenging to express concisely through natural language alone. These visual artifacts are widely employed across various domains, including engineering, software development, business process modelling, and education, due to their clarity and efficacy in illustrating relationships, sequences, and decision points. The inherent value of diagrams underscores the significance of enabling VLMs to comprehend their detailed content.

However, despite the general advancements in VLM capabilities, conventional models exhibit a critical shortcoming when it comes to understanding the fine-grained structural information present in diagram images. Structural elements, such as the directional arrows indicating process flow, lines representing connections, or symbols denoting branching conditions between nodes, carry immense semantic weight. Current VLMs often struggle to accurately capture these subtle yet crucial details~\cite{ye2025beyond}.

This deficiency is particularly problematic because even minor variations in a diagram's structural information can fundamentally alter its meaning. For instance, a single misplaced arrow, a swapped connection, or a slightly modified branching condition can completely change the interpretation of a process or system. The inability of VLMs to reliably discern these minute structural differences significantly hinders their utility in applications requiring precise diagrammatic understanding.

One straightforward approach to improving VLM performance on diagram understanding is to finetune the models on datasets of diagram images. However, this approach faces several challenges. Firstly, conventional contrastive learning techniques, commonly used for VLM finetuning, are often inadequate for distinguishing the subtle structural variations inherent in diagrams. These methods typically focus on learning broader semantic similarities and differences, failing to provide the granular distinction necessary for accurate diagrammatic interpretation. Secondly, many current VLMs do not support input with sufficient image resolution to comprehend the fine details of diagrams, which leads to the loss of this information through processes like image downscaling at the time of input.

The concept of hard negative sampling~\cite{patel2024tripletclip} has shown promise in improving contrastive learning by enabling models to learn finer distinctions between general images. Nevertheless, even with hard negative sampling, accurately identifying the precise structural information within diagram images remains a challenge. Furthermore, current hard negative sampling methods often rely on generative models for synthesising negative samples. These generative models typically struggle with the degree of precise control required to create subtle yet meaningful structural variations within diagrams that effectively highlight crucial differences.

Rule-based image editing has also been explored for generating hard negative samples and smaller sub-images that retain their resolution, demonstrating effectiveness in emphasising fine structural differences in diagrams for contrastive learning~\cite{sasaki2025structure}. However, this method, too, suffers from a significant constraint that necessitates the availability of component-editable image source files (e.g., vector graphics, hereafter referred to as ``editable data'') to perform the required image manipulations, thus limiting its applicability to real-world scenarios where such editable data are not available.

To overcome these limitations, we propose a novel contrastive learning framework that enhances the sensitivity of VLMs to fine-grained structural differences within diagram imagery without requiring editable data in its training dataset. Our method specifically addresses the challenge of VLMs acquiring of fine-grained structural awareness in diagrammatic information.

Our proposed technique involves, first, performing Optical Character Recognition (OCR) on a target diagram image to extract all textual labels associated with nodes and other elements. Subsequently, we generate synthetic pseudo diagram images by randomly constructing nodes and edges using a randomly selected subset of these extracted text labels. These pseudo diagrams are created alongside their corresponding editable data. This process allows us to create a diverse pool of structurally varied, albeit semantically nonsensical, diagrams. We then apply rule-based hard positive and negative sampling to these pseudo diagrams for contrastive learning. These randomly constructed diagrams, although semantically incoherent, retain sufficient structural variability to serve as effective hard positives and negatives.

We evaluated the model finetuned with our proposed methodology on flowchart images, a representative type of diagram. Our results demonstrate significant performance improvements in both text-image matching tasks and Visual Question Answering (VQA) tasks, regardless of the absence of semantically plausible training data, confirming the effectiveness of our approach in enhancing the VLM's ability to comprehend the intricate structural information within diagram images.

The key contributions of this work are:
\begin{itemize}
    \item We propose a novel method to generate hard positive and hard negative pairs for contrastive learning on diagrams, using only raster images without the need for editable formats (e.g., vector graphics) or pre-existing captions.
    \item Our method enables the creation of a finetuning dataset that emphasises subtle structural differences in diagrams while maintaining image sizes suitable for VLM inputs.
    \item We demonstrate through experiments on image-text matching and VQA tasks that finetuning with our method significantly improves a CLIP model's diagram understanding capabilities, even without access to editable data.
\end{itemize}

\section{Related Work}
This section reviews relevant literature in Vision-Language Models, diagram understanding, and contrastive learning, highlighting the context and motivation for our proposed approach.

\paragraph{Vision-Language Models}
Vision-Language Models (VLMs) are desined to bridge the semantic gap between visual and textual modalities. Pioneering works like CLIP~\cite{radford2021learning} and ALIGN~\cite{jia2021scaling} demonstrated the power of large-scale pre-training on image-text pairs, achieving impressive zero-shot generalisation capabilities across various tasks such as image classification, retrieval, and captioning. Subsequent models, including BLIP (Bootstrapping Language-Image Pretraining)~\cite{li2022blip} and Flamingo~\cite{alayrac2022flamingo}, further enhanced multimodal understanding by incorporating more sophisticated architectures and training strategies, leading to improved performance in tasks like visual question answering (VQA) and multimodal dialogue. These models excel at tasks involving natural images and general visual concepts, demonstrating robust performance in matching images to descriptive text or answering questions about common visual scenes. However, while these general-purpose VLMs exhibit strong performance on natural images, their inherent architecture and training data typically do not equip them with the specialised understanding required for highly structured visual information, such as diagrams.

\paragraph{Diagram specific vision models}
Whilst the considerable success of general-purpose VLMs like CLIP is well-demonstrated, their limitations in specialised contexts such as diagram interpretation have spurred the development of more tailored solutions.
One significant strand of research centres on finetuning pretrained models using diagram-specific datasets~\cite{masry2023unichart,zhang2025mavis}. For instance, UniChart~\cite{masry2023unichart} adapts the Donut architecture~\cite{kim2022ocr} for chart comprehension, finetuning it on a dataset of chart images. MAVIS~\cite{zhang2025mavis} introduces a mathmatical visual instruction tuning framework which leverages a custom vision encoder called CLIP-Math, which is a CLIP model finetuned on mathematical diagrams to facilitate visual problem solving.
Another strategy is diagram-specific preprocessing, such as the ``granulation'' method proposed in~\cite{sasaki2025structure}. This technique manipulates elements in vectorised images to subdivide large diagrams into smaller parts, enabling them to be input into VLMs without a loss of resolution.
These works suggest that task-specific data preparation strategies are crucial for enhancing the diagram understanding capabilities of VLMs.

\paragraph{Contrastive Learning and Hard Negative Sampling}
Contrastive learning has emerged as a highly effective paradigm for learning powerful representations by encouraging similar data points to be closer and dissimilar ones further apart in an embedding space. This approach has been central to the success of self-supervised learning methods in computer vision, including SimCLR~\cite{chen2020simple}, MoCo~\cite{he2020momentum}, and the aforementioned VLMs like CLIP. A critical component in the effectiveness of contrastive learning is the selection of negative samples. The quality and "hardness" of negative samples directly impact the model's ability to learn discriminative features. Hard negative mining strategies, which prioritise negative samples that are close to the anchor in the embedding space, have been shown to significantly boost performance by forcing the model to learn finer distinctions. Triplet loss and its variants, such as TripletCLIP~\cite{patel2024tripletclip}, explicitly leverage hard negative samples by constructing triplets of anchor, positive, and hard negative examples. Whilst these methods have demonstrated success in improving fine-grained discrimination for natural images, applying them effectively to diagram understanding presents unique challenges. Generating hard negative samples for diagrams that accurately reflect subtle structural differences, rather than just semantic content, is complex. Current generative models often struggle to maintain the precise structural integrity and semantic consistency required for creating meaningful hard negatives in this domain. Moreover, rule-based hard negative generation~\cite{sasaki2025structure}, while precise, frequently requires access to editable data, circumventing the core problem of raster-image-only understanding.

The reviewed literature highlights that existing methods are inadequate in robustly interpreting the subtle structural information within diagram imagery, especially when only raster image data is available. Our work addresses this critical limitation by proposing a novel contrastive learning framework that specifically targets the nuances of diagrammatic structure.

\section{Method}
\label{sec:method}
Our approach comprises three steps: editable pseudo diagram synthesis, rule-based hard contrastive data generation and structure-aware contrastive learning. The first step regenerates pseudo diagram images and captions in editable format from original (uneditable format) diagram images (detailed in \cref{sec:pseudo_synthesis}). In the subsequent step, a set of rule-based diagrammatic editing operations is applied to the editable pseudo diagram data to generate hard positive and negative samples (detailed in \cref{sec:contrastive_generation}). Lastly, model training is conducted via contrastive learning that attends the diagrammatic structual difference between the pseudo diagrams and their hard samples, as outlined in \cref{sec:sa_learning}.  
A visual summary of the process is provided in \cref{fig:overview}.

\subsection{Editable pseudo diagram synthesis}
\label{sec:pseudo_synthesis}
\begin{figure}[tb]
  \centering
   \includegraphics[width=1.05\linewidth]{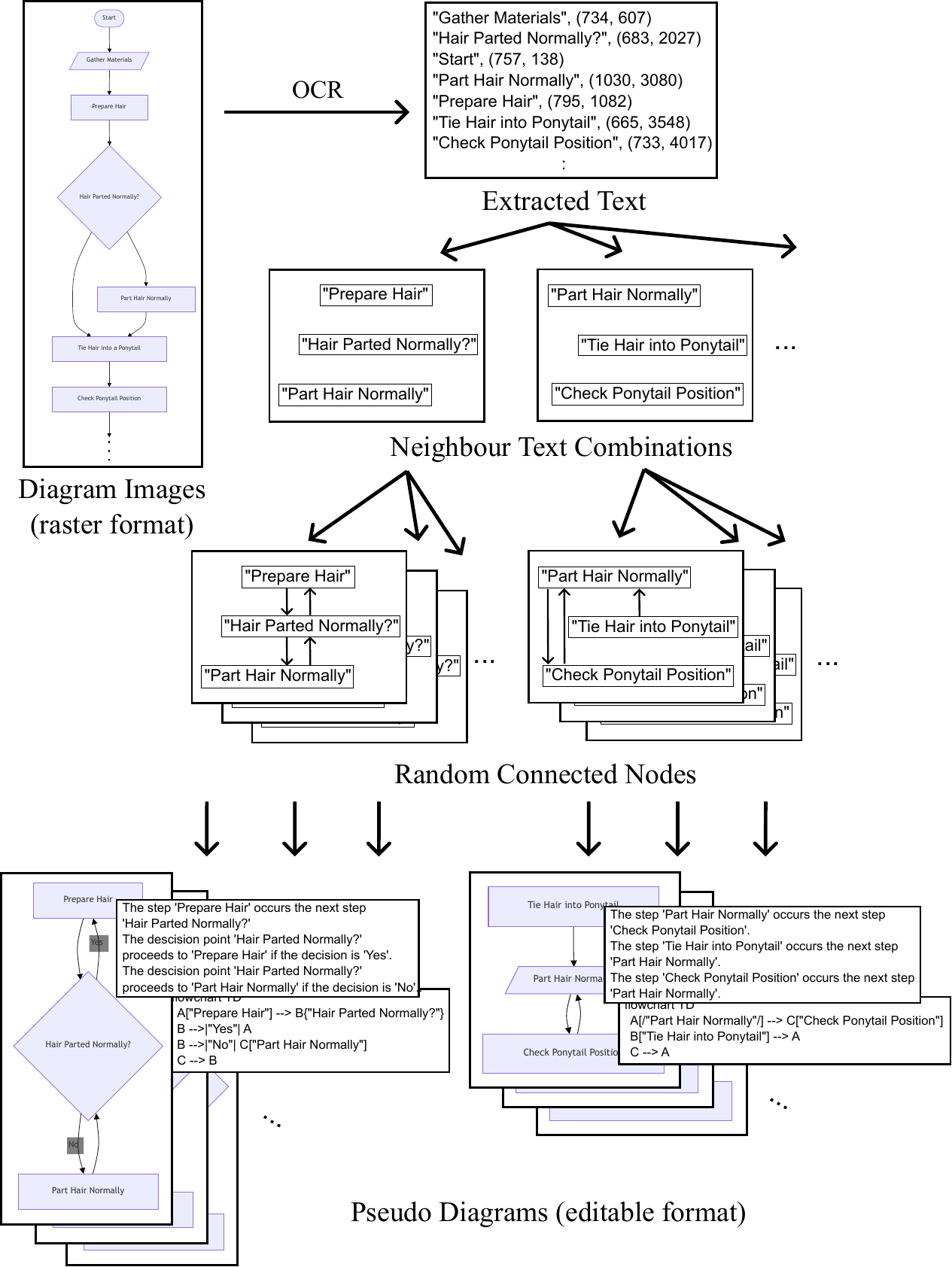}
  \caption{The editable pseudo diagram synthesis flow. OCR-extracted text is grouped by its proximity into combinations to form nodes. After adding random edge connections, this structure is finalised as definition code. This code is then used by a renderer to create both an editable image and a corresponding rule-based caption.}
   \label{fig:pseudo_diagrams}
\end{figure}

The pseudo diagram synthesis process is illustrated in \cref{fig:pseudo_diagrams}.
Let $R=\{r_i\}^{N^R}_{i=1}$ denote the set of original diagram images in raster-format. The process involves regenerating diagrams $D=\{d_{i,j}\}_{i=1:N^R,j=1:N^d}=\{(v_{i,j}, t_{i,j}, c_{i,j})\}_{i=1:N^R,j=1:N^d}$, where $v_{i,j}$ is the diagram image in an editable format such as vector graphics and $t_{i,j},c_{i,j}$ are the corresponding description and diagram definition codes (e.g., Mermaid code~\cite{std:mermaid}, referred to as ``diagram codes'' or simply ``codes''), by the steps in \cref{alg:pseuoe_synthesis}--\ref{alg:desc_generation}.
\begin{algorithm}[tb]
	\caption{Editable pseudo diagram synthesis}
	\label{alg:pseudo_synthesis}
	\renewcommand{\algorithmicrequire}{\textbf{Input}}
	\renewcommand{\algorithmicensure}{\textbf{Output}}
   \newcommand{\Desc}[2]{\Statex \makebox[6em][l]{#1:}#2}
   \algrenewcomment[1]{// #1}
   \algnewcommand{\LineComment}[1]{\State \color{comment} ~//#1\color{black}}
   \algnewcommand{\LineDesc}[2]{\makebox[5em][l]{#1}:~#2}
   \algnewcommand{\Initialize}[1]{\State {\bf Init}:~#1}
   \small
	\begin{algorithmic}[1]
		\Require $r_i$: Raster-format diagram image
        \Statex ~~~~$N^n$: Node size, $N^s$: Sampling size
		\Ensure $d_{i,j} $: Editable-format pseudo diagram data,
      \LineComment{Extract text}
      \LineComment{ $\textnormal{OCR}(\cdot)$: OCR operation}
      \LineComment{ $\{u^i_k\}^{N^o_i}_{k=1}=\{t^i_k,(x,y)^i_k\}^{N^o_i}_{k=1}$: Extracted text and positions}
        \For{$i \gets 1$ to $N^R$}
		    \State $\{u^i_k\}^{N^o_i}_{k=1} \gets \textnormal{OCR}(r_i)$;
        \EndFor
	  \LineComment{Select $N^s$ nearest neighbour text combinations as nodes}
      \LineComment{ $\textnormal{Dist}(c)$: Sum of the distances of text combinations $c$}
      \For{$i \gets 1$ to $N^R$}
        \State $C^i \gets \{\{u^i_{k_1},u^i_{k_2},\cdots,u^i_{k_{N^n}}\} | u^i_{k_m} \in \{u^i_k\}^{N^o_i}_{k=1}\}$;
        \State $\hat{C^i} \gets \textnormal{argmin}^{(N^s)}_{c^i \in C^i} \textnormal{Dist}(c^i)$;
      \EndFor
      \LineComment{Random connection}
      \LineComment{ $\textnormal{RC}(c)$: All combinations of}
      \LineComment{ ~~randomly connected directed links among nodes $c$}
      \For{$i \gets 1$ to $N^R$}
        \Initialize $G^i \gets \Phi$;
        \For{$c$ in $\hat{C^i}$}
          \For{$e$ in $\textnormal{RC}(c)$}
            \State $G^i \gets G^i \cup (c, e)$;
          \EndFor
        \EndFor
      \EndFor
      \LineComment{Generate diagrams}
      \LineComment{ $\textnormal{Code}(g)$: Code generate function}
      \LineComment{ $\textnormal{Desc}(g)$: Description generate function (Alg.~\ref{alg:desc_generation})}
      \LineComment{ $\textnormal{VG}(c)$: Vector-formatted diagram image renderer}
      \For{$i \gets 1$ to $N^R$}
        \Initialize $j \gets 1$;
        \For{$g$ in $G^i$}
          \State $d_{i,j} \gets (\textnormal{VG}(\textnormal{Code}(g)), \textnormal{Desc}(g), \textnormal{Code}(g))$;
          \State $j \gets j+1$;
        \EndFor
      \EndFor
	\end{algorithmic}
\end{algorithm}

\begin{algorithm}[tb]
	\caption{Diagram description generate function}
	\label{alg:desc_generation}
	\renewcommand{\algorithmicrequire}{\textbf{Input}}
	\renewcommand{\algorithmicensure}{\textbf{Output}}
   \newcommand{\Desc}[2]{\Statex \makebox[6em][l]{#1:}#2}
   \algrenewcomment[1]{// #1}
   \algnewcommand{\LineComment}[1]{\State \color{comment} ~//#1\color{black}}
   \algnewcommand{\LineDesc}[2]{\makebox[5em][l]{#1}:~#2}
   \algnewcommand{\Initialize}[1]{\State {\bf Init}:~#1}

   \small
	\begin{algorithmic}[1]
		\Require $g=(c,e)$: Diagram nodes ($c=\{c_i\}$) and
      \Statex ~~~~~connections ($e=\{(i^\textnormal{from}_j, i^\textnormal{to}_j)\}$)
		\Ensure $t=\{t_j\}$: Diagram description,
        \For{$i \gets 1$ to $N^c$}
          \Initialize $c^d_i \gets$ ``Yes'';
        \EndFor
        \For{$j \gets 1$ to $N^e$}
          \State $t_j \gets$ ``From $c_{i^\textnormal{from}_j}$: '';
            \If{$c_{i^\textnormal{from}_j}$ is question.}
		      \State $t_j \gets t_j +$ ``If **'' $+ c^d_i +$ ``**, proceed to '';
              \If{$c^d_i$ is ``Yes''}
                \State $c^d_i \gets$ ``No'';
              \Else
                \State $c^d_i \gets$ ``'';
              \EndIf
            \Else
              \State $t_j \gets t_j +$ ``Proceed to '';
            \EndIf
            \State $t_j \gets t_j + c_{i^\textnormal{to}_j}$;
        \EndFor
	\end{algorithmic}
\end{algorithm}

\subsection{Rule-based hard contrastive data generation}
\label{sec:contrastive_generation}
\begin{figure}[t]
    \centering
     \includegraphics[width=1.0\linewidth]{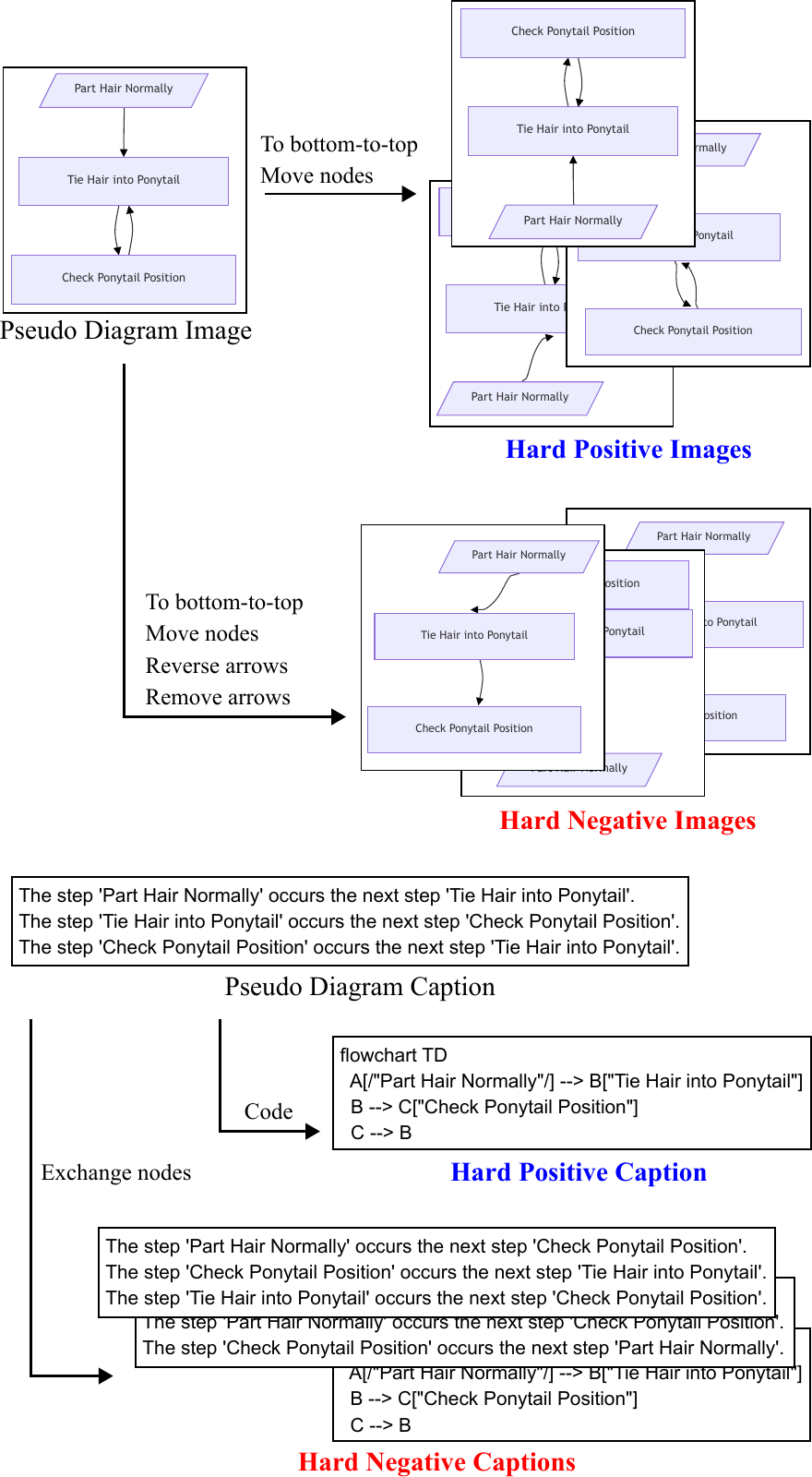}
    \caption{The synthesis of hard positive and negative samples. Hard positive images are generated via minor visual edits, such as altering node positions or flow orientation. Hard negatives are created by permuting nodes and edges. The original diagram code serves as the hard positive caption, while hard negative captions are generated by swapping labels within the code to create a structural mismatch.}
     \label{fig:hard_sample}
  \end{figure}

A set of the image editing operations is applied to the pseudo diagram data obtained by the steps in \cref{sec:pseudo_synthesis}, as proposed in~\cite{sasaki2025structure} but we modify the rules which outlined below:
\begin{description}
    \item[Rule for Hard Positive Images] \hfill \\
    Apply a combination of the following editions:
    \begin{itemize}
        \item Randomly reverse the flow direction from top-down to bottom-up (since all original diagrams are defined with a top-down flow). 
        \item Randomly move the positions of selected nodes.
    \end{itemize} 
    \item[Rule for Hard Negative Images] \hfill \\
    Apply a combination of the following perturbations:
    \begin{itemize}
        \item Reverse the direction of randomly selected arrows.
        \item Remove a subset of arrows at random.
        \item Apply ``Rule for Hard Positive Image'' after the above operations.
    \end{itemize} 
    \item[Rule for Hard Positive Captions (same as~\cite{sasaki2025structure})] \hfill \\
        The code of the diagram. 
    \item[Rule for Hard Negative Captions (same as~\cite{sasaki2025structure})] \hfill \\
    Apply a combination of the following semantic distortions:
    \begin{itemize}
        \item Randomly swap the labels of selected nodes in the natural language description. \hfill 
        \item Randomly swap the labels of nodes within the codes.
    \end{itemize} 
\end{description}
As shown in~\cref{fig:hard_sample}, hard positive and negative samples are generated by applying these editing rules.

\subsection{Structure-aware contrastive learning}
\label{sec:sa_learning}
The diagram data and their hard positive / negative samples prepared in \cref{sec:pseudo_synthesis} and \cref{sec:contrastive_generation} are used for contrastive learning specialised to diagrammatic imagery. We use structure-aware contrastive learning~\cite{sasaki2025structure}, which has been reported its performance strength in a diagram image domain. The objective is defined as:
\begin{align}
    \mathcal{L} = \mathcal{L}_\textnormal{CL}+\lambda_\textnormal{SC}\mathcal{L}_\textnormal{SC}+\lambda_\textnormal{DO}\mathcal{L}_\textnormal{DO},
    \label{eqn:total_loss}
\end{align}
where $\mathcal{L}_\textnormal{CL}$, $\mathcal{L}_\textnormal{SC}$ and $\mathcal{L}_\textnormal{DO}$ are the InfoNCE loss~\cite{oord2018representation}, structure-aware contrastive (SC) loss and distinct factor orthgonal (DO) loss, respectively. $\lambda_\textnormal{SC}$ and $\lambda_\textnormal{DO}$ are hyperparameters controlling the relative contributions of the SC and DO losses. We omit the DO loss ($\lambda_\textnormal{DO}=0$) for model training in this paper since no significant differences in the performance are reported in \cite{sasaki2025structure}.

The InfoNCE loss is defined as:
\begin{align}
    \mathcal{L}_\textnormal{CL} = \frac{1}{2} \mathbb{E}_{v_i, t_i}\left[ -\log \frac{\exp (\textnormal{sim} (F_v(v_i), F_t(t_i) ) )}{\mathbb{E}_{v_i, t_{j\neq i}}[\langle F_v(v_i), F_t(t_j) \rangle]} \right. \nonumber \\
    \left. - \log \frac{\exp \langle F_v(v_i), F_t(t_i) \rangle}{\mathbb{E}_{v_{j\neq i}, t_i}[\langle F_v(v_j), F_t(t_i) \rangle]}\right] ,
    \label{eqn:clip_loss}
\end{align}
where $\textnormal{sim}(\cdot, \cdot)$ denotes the similarity function, originally defined as cosine similarity, $v_i$, $t_i$, $F_v$ and $F_t$ represent the training image data, their captions, the vision and text encoder, respectively.
For clarity, feature normalization and the temperature parameter are omitted in the equations in this paper. We adopt an absolute value of the cosine similarity for the similarity function, since we find this modification contributes the stability of the model training of our method, as:
\begin{align}
    \textnormal{sim} (F_v(v_i), F_t(t_i) ) = \left| \langle F_v(v_i), F_t(t_i) \rangle \right|.
    \label{eqn:similarity}
\end{align}

The SC loss is calculated using the positive similarity $\mathcal{S}^p(v_i, t_i)$ and negative similarity $\mathcal{S}^n(v_i, t_i)$ between samples, which are defined as follows:
\begin{align}
    \mathcal{S}^p&(v_i, t_i) = \nonumber \\
    &\mathbb{E}_{v^p\sim V^p(v_i), t^p\sim T^p(t_i)} [\exp ( \textnormal{sim} (F_v(v_i), F_v(v^p)) ) + \nonumber \\
    &~\exp ( \textnormal{sim} (F_t(t_i), F_t(t^p)) ) + \exp ( \textnormal{sim} (F_v(v_i), F_t(t^p)) ) + \nonumber \\
    &~\exp ( \textnormal{sim} (F_t(t_i), F_v(v^p)) ) ],
    \label{eqn:positive_similarity}
\end{align}
\vspace{-4mm}
\begin{align}
    \mathcal{S}^n&(v_i, t_i) = \nonumber \\
    &\mathbb{E}_{v^n\sim V^n(v_i), t^n\sim T^n(t_i)} [\exp ( \textnormal{sim} (F_v(v_i), F_v(v^n)) ) + \nonumber \\
    &~\exp ( \textnormal{sim} (F_t(t_i), F_t(t^n)) ) + \exp ( \textnormal{sim} (F_v(v_i), F_t(t^n)) ) + \nonumber \\
    &~\exp ( \textnormal{sim} (F_t(t_i), F_v(v^n)) )],
    \label{eqn:negative_similarity}
\end{align}
where $V^p(\cdot), T^p(\cdot), V^n(\cdot), V^n(\cdot)$ are the generate functions for hard positive images, hard positive captions, hard negative images, and hard negative captions, respectively.
The SC loss is then formulated as:
\begin{align}
    \mathcal{L}_\textnormal{SC} = \mathbb{E}_{v_i, t_i} \left[ -\log \frac{\mathcal{S}^p(v_i, t_i)}{\mathcal{S}^p(v_i, t_i)+\mathcal{S}^n(v_i, t_i)} \right].
    \label{eqn:rcloss}
\end{align}
This loss encourages the model to minimise the distance between the original and hard positive samples, while maximising the distance between the original and hard negative samples.

\section{Evaluation}
\label{sec:evaluation}
We evaluate the proposed method in \cref{sec:method} on flowcharts, a representative example of diagrams.

\subsection{Baselines}
\label{sec:baselines}
The evaluation compares our method against the following cases:
\begin{description}
    \item[w/o FT (zero-shot baseline)] The publicly available CLIP ViT-L/14@336px model~\cite{clip:vit-large-patch14} without any task-specific finetuning.
    \item[CROP] Finetuned the CLIP ViT-L/14@336px model with the image cropping operation applied to the input of the model.
    \item[GRAN] Finetuned the CLIP ViT-L/14@336px model with the granulated and hard samples generated in the diagrammatic editing method~\cite{sasaki2025structure}, assuming the easy case that editable data of flowchart images are available. The objective functions for the model optimisation are varied by using either the CLIP loss~\cref{{eqn:clip_loss}} (CLIP) or the SC loss~\cref{eqn:rcloss} (SaCLIP).
    \item[PSEUDO (ours)] Finetuned the CLIP ViT-L/14@336px model with the reconstructed data with hard samples via our method (see~\cref{sec:method}), which generates editable format of diagrammatic data with hard samples, assuming the case that only raster format of flowchart images are available. The objective functions for the model optimisation are varied by using either the CLIP loss~\cref{{eqn:clip_loss}} (CLIP) or the SC loss~\cref{eqn:rcloss} (SaCLIP).
\end{description}

\subsection{Dataset Preparation}
\label{sec:dataset}
For the experiments, we utilise the publicly available FlowVQA dataset~\cite{singh2024flowvqa}, which includes a set of flowchart images, corresponding textual descriptions, Mermaid code representations, and QA pairs. We use mermaid-cli~\cite{code:mermaid-cli-9-4-0} as a diagram image renderer throughout this work.

\paragraph{Training data}
For the case of {\bf CROP}, each disgram image is randomly cropped to the input size of the CLIP model 59 times, which is to make the total number of instances comparable to that of the GRAN case, resulting the number is 38,765. The cropped images are stored as PNG~\cite{std:png} files. The caption for each cropped image is defined by concatenating the words extracted by Tesseract OCR~\cite{smith2007overview,code:tesseract-5-1-1}.

For the case of {\bf GRAN}, the Mermaid codes are parsed and transformed according to the granulation procedure in~\cite{sasaki2025structure}. This process involves extracting all combinations of adjacent triplets of nodes from each diagram code to regenerate simplified versions of the original codes. The image captions are generated from the definition of the nodes and connections in the Mermaid codes via \cref{alg:desc_generation}. The total number of the granulated instances is 37,662.
The resulting flowchart images are rendered as SVG~\cite{std:svg} files to be edit for hard positive and negative sample synthesis.

For the case of {\bf PSEUDO}, only images in the flowchart dataset are used to constract the training data. The reconstructed (editable) data is generated by the process in~\cref{sec:pseudo_synthesis}. The node size $N^n$ is set to 3, as the same as the granulation process. We also use Tesseract OCR for the text extraction in the process. The total number of the reconstructed pseudo diagrams is controlled by random sampling to match the size of the GRAN case. The flowchart images are rendered in the same manner as the GRAN case.

\vspace{-2mm}
\paragraph{Test data}
The test data is prepared by applying the granulation process to the test set of the FlowVQA dataset as in~\cite{sasaki2025structure}.

\begin{table*}[tb]
    \begin{center}
        \caption{Evaluation results for the image-text matching task, reporting Recall@1 (R@1), Recall@5 (R@5), Recall@10 (R@10), and Mean Reciprocal Rank (MRR) across different data preparation and finetuning strategies.}
        \begin{tabular}{l c c c c c c c c} \hline
                    & \multicolumn{4}{c}{Query: images, Retrieve: captions} & \multicolumn{4}{c}{Query: captions, Retrieve: images} \\
                   & R@1 & R@5 & R@10 & MRR & R@1 & R@5 & R@10 & MRR \\ \hline
            w/o FT~\cite{clip:vit-large-patch14} & 0.65726 & 0.88790 & 0.92231 & 0.75842 & 0.66586 & 0.89812 & 0.93575 & 0.76584 \\
            CROP - CLIP~\cite{radford2021learning} & 0.48118 & 0.70161 & 0.77419 & 0.57887 & 0.48118 & 0.65081 & 0.70726 & 0.55689 \\
            GRAN~\cite{sasaki2025structure} - CLIP~\cite{radford2021learning} & 0.76694 & 0.92850 & 0.94758 & 0.84036 & 0.77930 & {\bf 0.93925} & 0.94893 & 0.85005 \\
            GRAN~\cite{sasaki2025structure} - SaCLIP~\cite{sasaki2025structure} & 0.76882 & 0.92796 & 0.94785 & 0.84148 & 0.77903 & 0.93253 & 0.94570 & 0.84869 \\
            PSEUDO (ours) - CLIP~\cite{radford2021learning} & 0.76667 & {\bf 0.93952} & 0.95645 & 0.84315 & 0.78253 & 0.93790 & {\bf 0.95591} & 0.85319 \\
            PSEUDO (ours) - SaCLIP~\cite{sasaki2025structure} & {\bf 0.77097} & {\bf 0.93952} & {\bf 0.95914} & {\bf 0.84542} & {\bf 0.78763} & 0.93871 & 0.95484 & {\bf 0.85532} \\ \hline
        \end{tabular}
        \label{tab:ranking}
    \end{center}
\end{table*}
\begin{table*}[tb]
    \vspace{-1mm}
    \begin{center}
        \caption{Evaluation results for the image-text matching task in the presence of hard negative samples, reported using Recall@1 (R@1), Recall@3 (R@3), and Mean Reciprocal Rank (MRR) across different data preparation and finetuning strategies.}
        \begin{tabular}{l c c c c c c} \hline
                    & \multicolumn{3}{c}{Query: images, Retrieve: captions} & \multicolumn{3}{c}{Query: captions, Retrieve: images} \\
                   & R@1 & R@5 & MRR & R@1 & R@5 & MRR \\ \hline
            w/o FT~\cite{clip:vit-large-patch14} & 0.16613 & 0.16613 & 0.16613 & 0.60538 & 0.82419 & 0.71632 \\
            CROP - CLIP~\cite{radford2021learning} & 0.20887 & 0.20887 & 0.20887 & 0.49140 & 0.75403 & 0.62629 \\
            GRAN~\cite{sasaki2025structure} - CLIP~\cite{radford2021learning} & 0.46667 & 0.46667 & 0.46667 & 0.61640 & 0.82634 & 0.72529 \\
            GRAN~\cite{sasaki2025structure} - SaCLIP~\cite{sasaki2025structure} & {\bf 0.47527} & {\bf 0.47527} & {\bf 0.47527} & {\bf 0.89973} & {\bf 0.98602} & {\bf 0.94133} \\
            PSEUDO (ours) - CLIP~\cite{radford2021learning} & 0.35269 & 0.35269 & 0.35269 & 0.50511 & 0.76022 & 0.63644 \\
            PSEUDO (ours) - SaCLIP~\cite{sasaki2025structure} & 0.38817 & 0.38817 & 0.38817 & 0.70484 & 0.87204 & 0.79069 \\ \hline
        \end{tabular}
        \label{tab:ranking_hard}
    \end{center}
\end{table*}

\vspace{-2mm}
\paragraph{Hard sample synthesis}
We employ the hard sample synthesis method from~\cite{sasaki2025structure}. In accordance with~\cref{sec:contrastive_generation}, rendered SVG images and Mermaid codes are used to create hard positive and negative pairs, consisting of both images and captions.
During training, this process is applied on-the-fly to each batch, fostering robustness by ensuring constant exposure to challenging examples.
In contrast, for evaluation, we pre-generate a static set of hard negatives for every test set sample. This fixed set, which includes six captions and eight images, guarantees that all models are compared under identical conditions, ensuring fair and reproducible results.

\subsection{CLIP Training}
\label{sec:clip_training}
The finetuning process in \cref{sec:baselines} is conducted using the dataset introduced in \cref{sec:dataset}. We employ the LoRA method~\cite{hu2022lora}, applying it to all linear layers with the parameters set to $\alpha=16$ and $r=8$. All images are converted to a raster format prior to being input into the model. The training is carried out for 3 epochs with a learning rate of 0.0004, following a warmup period of 8 steps. A batch size of 32 is used with 4 gradient accumulation steps, creating an effective batch size of 128. For our proposed method, the hyperparameter $\lambda_\textnormal{SC}$ was fixed at 0.1, in line with the value used in~\cite{sasaki2025structure}.

\subsection{Image-text Matching Task Evaluation}
The setup for the image-text matching task follows~\cite{sasaki2025structure}.
All images and captions from the test set described in~\cref{sec:dataset} are encoded into feature representations using the respective encoders of the models finetuned as in~\cref{sec:baselines}. For each image, the top 10 most similar captions are retrieved, and vice versa, based on cosine similarity. To evaluate image-text matching accuracy, we compute Recall@1 (R@1), Recall@5 (R@5), Recall@10 (R@10), and Mean Reciprocal Rank (MRR) over the retrieved sets. The results are presented in~\cref{tab:ranking}.

Our proposed method achieves notable improvements except for R@5 for image retrieval. Note that our method does not need editable data unlike other diagram-specific finetuning methods.

To further assess robustness, we conduct the same experiment under more challenging conditions involving hard negatives. Specifically, for each image, the top 3 most similar captions are retrieved from a candidate set comprising the correct caption and hard negatives; the same process is applied for each caption. We report R@1, R@3, and MRR in~\cref{tab:ranking_hard}. The results show the degradation in performance for our method compared to a conventional diagram-specific contrastive learning framework that requires editable data; however, they also show that our method outperforms all of other cases in which editable data is not available in this harder setting.

\subsection{VQA Task Evaluation}
\begin{table*}[tb]
    \begin{center}
        \caption{A comparison of the BERTScore~\cite{zhang2020bertscore} performance across different LLaVA~\cite{liu2024visual} models with their vision encoders swapped.}
        \begin{tabular}{l c c c} \hline
                  Vision Encoder & Precision & Recall & F1 \\ \hline
                w/o FT & 0.483461 & 0.425924 & 0.494744 \\
                CROP - CLIP~\cite{radford2021learning} & {\bf 0.713844} & 0.543008 & 0.618101 \\
                GRAN~\cite{sasaki2025structure} - CLIP~\cite{radford2021learning} & 0.532666 & 0.467946 & 0.507943 \\
                GRAN~\cite{sasaki2025structure} - SaCLIP~\cite{sasaki2025structure} & 0.620144 & 0.601086 & 0.634233 \\
                PSEUDO - CLIP (ours) & 0.558813 & 0.488994 & 0.536966 \\
                PSEUDO - SaCLIP (ours) & 0.660028 & {\bf 0.612537} & {\bf 0.647963} \\ \hline
        \end{tabular}
        \label{tab:llm_results}
    \end{center}
\end{table*}
To assess the downstream impact of our work, we evaluate our finetuned CLIP model within a VQA setting, in line with the approach taken by~\cite{sasaki2025structure}. The experiment involves integrating our model as a vision encoder into a CLIP-based large language model (LLM). We selected LLaVA-v1.6-Mistral-7B~\cite{llava:v1.6-mistral-7b} for this task, as this widely used LLaVA~\cite{liu2024visual} variant features a CLIP ViT-L/14@336px encoder, which shares the same architecture as our model and thus provides a suitable test environment.

This evaluation uses the original FlowVQA dataset, departing from the granulated versions used previously. The LLM's ability to process higher-resolution inputs through a split-and-merge technique~\cite{liu2024improved} makes this feasible. The core of the experiment is a comparison of performance, wherein we replace the LLM's default vision encoder with each of the model variants described in~\cref{sec:baselines}.

\paragraph{LLaVA Training}
Following the procedure in~\cite{sasaki2025structure}, we finetuned all modules except for the vision encoders. For this, we employed the LoRA method~\cite{hu2022lora} on all linear layers, configured with $\alpha=32$ and $r=16$. The models were trained for 3 epochs using a learning rate of 0.0004 after an 8-step warmup. We used a batch size of 2 and 16 gradient accumulation steps, resulting in an effective batch size of 32.

\paragraph{VQA Performance}
The setup for this task follows~\cite{sasaki2025structure}. We assess model performance on the FlowVQA test set using the average BERTScore~\cite{zhang2020bertscore} to measure the similarity between model-generated answers and the ground truth. As presented in \cref{tab:llm_results}, the findings demonstrate that finetuning the vision encoder substantially enhances performance. Our method, in particular, outperforms the baseline CLIP finetuning approaches, achieving higher recall and F1 scores. A key advantage of our technique is that it obviates the need for curated image-caption pairs, instead generating them solely from the raster images.

\subsection{Limitation} 
Whilst the experiments presented in this paper demonstrate the effectiveness of our method in improving performance on flowchart image-text matching and VQA tasks, a notable limitation regarding the dataset assumption remains. Our method heavily relies on the assumption of the similar appearance between diagrams in a target domain and the output of a diagram renderer. It is not fully applicable when the appearance of diagrams in a target domain is different from a style that the diagram renderer does not support. Furthermore, the quality of the pseudo-diagrams is degraded when the type of the target diagram is not supported by the diagram renderer. This limites the range of application of our approach.

Another limitation is possibly arises from our reliance on OCR; the impact of the quality of the OCR output, which is currently unclear.  

Further work is required to explore effectiveness of our approach in diverse diagrammatic contexts and impact of OCR quality.

\section{Conclusion}

In this paper, we address a critical limitation of conventional Vision-Language Models (VLMs) about inability to comprehend the fine-grained structural details within diagram imagery. We identify that existing finetuning approaches are often hampered by an inability to distinguish subtle yet meaningful structural variations, or by a reliance on editable source data, such as vector graphics, which are frequently unavailable.

To overcome these challenges, we introduce a novel contrastive learning framework that enables VLMs to acquire a nuanced understanding of diagrammatic structures using only raster images. Our key innovation is a method for synthesising a diverse set of pseudo diagrams and their corresponding editable data from a single source image. This allows for the application of rule-based hard negative and hard positive sampling to generate a training dataset that effectively emphasises fine structural differences, without requiring any pre-existing editable files or semantically plausible training examples.

Our experimental evaluation on flowchart-based image-text matching and Visual Question Answering tasks demonstrate that a CLIP model finetuned with our methodology achieves significant performance improvements. These results confirm that our approach successfully enhances the model's sensitivity to crucial structural information, proving that VLM's diagrammatic reasoning can be effectively improved even in the absence of manually curated or editable diagram datasets.

The proposed framework broadens the applicability of finetuning VLMs for specialised domains where data is limited to non-editable formats. Future work could explore extending this method to a wider variety of diagram types and investigating its resilience to OCR inaccuracies. Nevertheless, this work represents a significant step towards developing more practical and robust VLMs capable of deep and accurate diagram comprehension.
{
    \small
    \bibliographystyle{ieeenat_fullname}
    \bibliography{references}
}

\end{document}